\documentclass[letterpaper, 10 pt, conference]{ieeeconf}

\IEEEoverridecommandlockouts
\usepackage[english]{babel}
\usepackage{amsmath}
\usepackage{amssymb}
\usepackage{booktabs}
\usepackage{caption}
\usepackage{subfigure}
\usepackage[dvips]{graphicx}
\usepackage{multirow}
\usepackage{amsfonts}
\usepackage{breakurl}
\usepackage{enumerate}
\usepackage{bm}
\usepackage{enumerate}
\usepackage{adjustbox}
\usepackage{siunitx}
\usepackage{booktabs}
\usepackage{mdframed}
\usepackage{adjustbox}
\usepackage{authblk}
\usepackage{color}
\usepackage{xurl}
\usepackage{cite}
\usepackage{cite}
\makeatletter
\let\NAT@parse\undefined
\makeatother
\usepackage{hyperref}

\usepackage[dvipsnames, table]{xcolor}
\usepackage[T1]{fontenc}
\usepackage{amsmath,amssymb}
\usepackage{scalerel}
\usepackage{siunitx}
\usepackage{comment}
\usepackage{graphicx}
\usepackage{subfigure}
\usepackage{multirow}
\usepackage[ruled,vlined]{algorithm2e}
\usepackage{bm}
\usepackage{url}
\usepackage{hyperref}
\usepackage{diagbox} 
\usepackage{balance}
\usepackage{utfsym}
\usepackage{pifont}
\usepackage{subfigure}
\usepackage{eso-pic}
\newcommand\AtPageUpperMyright[1]{\AtPageUpperLeft{%
		\put(\LenToUnit{2cm},\LenToUnit{-1cm}){%
			\parbox{\textwidth}{\centering\fontsize{9}{11}\selectfont #1}}%
}}%
\newcommand{\conf}[1]{%
	\AddToShipoutPictureBG*{%
		\AtPageUpperMyright{#1}
	}
}
\newcommand{\cmark}{\text{\ding{51}}}
\newcommand{\xmark}{\text{\ding{55}}}

\newcommand{\ie}{\textit{i}.\textit{e}., }
\newcommand{\eg}{\textit{e}.\textit{g}., }

\newcommand\changes[1]{\textcolor{black}{#1}}
\newcommand\changesIROS[1]{\textcolor{black}{#1}}

\begin{document}

\title{\LARGE \bf The Power of Input: Benchmarking Zero-Shot Sim-to-Real Transfer of Reinforcement Learning Control Policies for Quadrotor Control}

\author{Alberto~Dionigi$^{1,2}$, Gabriele~Costante$^{2}$, and Giuseppe~Loianno$^{1}$
\thanks{$^{1}$The authors are with the New York University, Tandon School of Engineering, Brooklyn, NY 11201 USA. {\tt\footnotesize loiannog@nyu.edu}.}
\thanks{$^{2}$The authors are with the Department of Engineering, University of Perugia, 06125 Perugia, Italy {\tt\footnotesize alberto.dionigi@unipg.it, gabriele.costante@unipg.it}.}
\thanks{This work was supported by the NSF CAREER Award 2145277, the DARPA YFA Grant D22AP00156-00, Qualcomm Research, Nokia, and NYU Wireless. G. Costante acknowledges funding by PNRR-M4C2 – I1.1 – MUR Call for proposals n. 1409 del 14-09-2022 - Bando PRIN 2022
PNRR - ERC sector PE6- Project title: LiSA - Listen, See and Act: fusing audio-video cues 
to perceive visible and invisible events and develop perception-to-action solutions for 
autonomous vehicles - Project Code P2022MEBFM- CUP Code D53D23017510001 (CUP Code of the University of Perugia Unit: J53D23015010001) -
Funded by the European Union – NextGenerationEU.}
}

\thispagestyle{empty}
\pagestyle{empty}
\makeatletter
\g@addto@macro\@maketitle{
    \vspace{10pt}
    \setcounter{figure}{0}
    \centering
    \includegraphics[width=\linewidth]
    {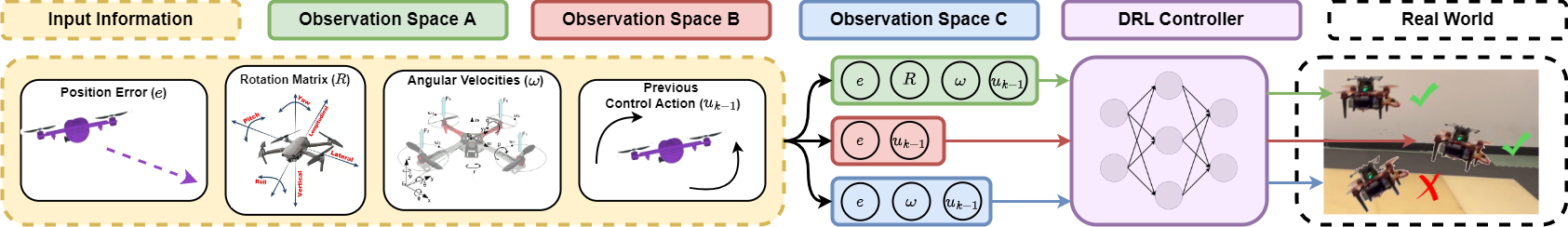}
    \captionof{figure}{\changes{Overview of the benchmark study: we investigate the impact that different input configurations have on the ability of Deep Reinforcement Learning control policies to fly a real drone with zero-shot sim-to-real adaptation.}
    \label{fig:overview}}
    \vspace{-10pt}
}
\makeatother
\maketitle

\conf{This work has been accepted to the IEEE/RSJ International Conference on Intelligent Robots and Systems (IROS). This is an archival version of our paper. Please cite the published version DOI: \url{https://doi.org/10.1109/IROS58592.2024.10802831}}

\begin{abstract}
In the last decade, data-driven approaches have become popular choices for quadrotor control, thanks to their ability to facilitate the adaptation to unknown or uncertain flight conditions. Among the different data-driven paradigms, Deep Reinforcement Learning (DRL) is currently one of the most explored. However, the design of DRL agents for Micro Aerial Vehicles (MAVs) remains an open challenge. While some works have studied the output configuration of these agents (\ie what kind of control to compute), there is no general consensus on the type of input data these approaches should employ. Multiple works simply provide the DRL agent with full state information, without questioning if this might be redundant and unnecessarily complicate the learning process, or pose superfluous constraints on the availability of such information in real platforms. In this work, we provide an in-depth benchmark analysis of different configurations of the observation space. We optimize multiple DRL agents in simulated environments with different input choices and study their robustness and their sim-to-real transfer capabilities with zero-shot adaptation. We believe that the outcomes and discussions presented in this work supported by extensive experimental results could be an important milestone in guiding future research on the development of DRL agents for aerial robot tasks.
\end{abstract}
\IEEEpeerreviewmaketitle

\section{Introduction}
\label{sec:intro}

Recent years have seen the emergence of MAVs platforms like quadrotors that has revolutionized the research and commercial fields of robotics. Their agility and maneuverability make them suitable to be effectively deployed in several application scenarios including, but not limited to surveillance, environmental monitoring, and search and rescue \cite{emran2018reviewquadrotor}. 

In the last decades, quadrotor control solutions have primarily relied on model-based \cite{voos2009nonlinear} and model-free approaches \cite{younes2016robust} stemming from control system theory. However, recently data-driven methods have shown their ability to infer the control policies from data and experience and are able to directly map sensor readings to control actions. This relaxes or eliminates the need for modular architectures (\eg a controller coupled with a state estimator), or knowledge about the system dynamics \cite{pfeiffer2022visual} consequently reducing the control latency and potentially increasing the system adaptation to multiple flight conditions. Among these, the Reinforcement Learning (RL) paradigm is certainly one of the most promising strategies to enable the design of end-to-end controllers.

However, existing works on RL-based control of aerial robots propose different model designs without a general consensus on which is the most suitable input set that is sufficient for the policy network to guarantee effective control. Indeed, the RL control agents proposed in literature are designed to process observations of the full system state, indirectly assuming that more information leads to better performance\cite{wang2019deterministic, song2021autonomous}. This hypothesis might not hold for two main reasons: i) some observations might not be available on the real drone platform; ii) providing more information does not necessarily ease the optimization of the RL agent, which might instead be more complex due to the increased dimension and redundant information of the input space.

Motivated by these considerations, in this work, we conduct an in-depth analysis on multiple possible information choices provided to the RL agent to compute the control policy for quadrotor control. For this purpose, different input modalities are considered and several RL agents are trained in simulation. Evaluation, on the other hand, is performed by deploying these models (without any fine-tuning) on real drone platforms. This allows to study the robustness of the different input modalities to non-ideal conditions not modeled in simulation. More broadly, this allows to assess the sim-to-real capabilities of the different models providing to the community an analysis that can be directly exploited to design drone controllers for practical applications. Understanding the effect that different input configurations have on the performance and on the sim-to-real transfer capabilities of RL-based drone controllers is crucial to design robust and reliable systems. Is it more convenient to provide the policy network with all the available information on the robot state (\eg position and orientation in the real-world fixed frame, angular velocities, and accelerations), leaving to the RL algorithm the role of combining and mapping \changes{data} to control actions? Or is providing additional much (and possibly redundant) information harmful to the RL optimization process? 

These questions are still unanswered and, in general, an analysis of the impact of the input data on the RL-based controller performance is still missing. Therefore, this paper presents the following key contributions
\begin{itemize}
    \item We propose the first benchmark analysis of RL-based control policy for aerial robots focused on input configurations. We consider multiple possible design choices optimizing in simulation environments different models that process different data types ranging from minimal sensor readings to the full drone state information.
    \item We present an extensive experimental campaign to study the performance of different RL agents and compare their sim-to-real transfer capabilities. To achieve this, the controllers \changes{are trained in simulation on a randomized family of MAV dynamic models, and}
    are deployed without any fine-tuning on a real quadrotor platform, presenting characteristics that inevitably differ from those of the simulated systems. This approach allows for accurate comparison of the models' performance in real-world scenarios, which is crucial for their practical applications.
    \item \changesIROS{We highlight that increasing the information provided to the DRL agent does not necessarily enhance the performance. Furthermore, we show that configurations with a limited observation space (\eg containing only relative position data) are able to learn robust flying policies and can achieve comparable performance with respect to models with more privileged information.}
\end{itemize}


\changes{The rest of the paper is organized as follows. Section \ref{sec:related} summarizes the state-of-the-art on control strategies for quadrotors, highlighting that nearly no works have studied the design choices for successfully sim-to-real transfer of the RL agents. Subsequently, Section \ref{sec:method} introduces the proposed methodology, while Section \ref{sec:experiments} presents several experimental tests and discusses the results. Finally, Section \ref{sec:conclusion} concludes the paper and reports possible future research directions.}

\section{Related Work}
\label{sec:related}


\textbf{Classic Control Methods}. Stabilization and position control of aerial platforms have gathered significant attention in literature, leading to the development of multiple classic control techniques \cite{kim2019comprehensive}. These methods encompass both model-free approaches, such as proportional-integral-derivative (PID) controllers, and model-based strategies, including Linear Quadratic Regulator (LQR) and Model Predictive Control (MPC). The PID stands out as the most popular choice due to its ability to achieve good set-point stabilization performance \cite{bouabdallah2004design, bouabdallah2004pid} with minimal implementation effort. 
However, since aerial platforms are highly non-linear and under-actuated systems \cite{kim2016comparison}, the performance of PID controllers considerably decreases in more challenging scenarios where agile flight with low position error is required. 

More advanced solutions that exploit the knowledge of the dynamic model have, therefore, emerged. Among those,  feedback linearization approaches \cite{fritsch2012quasi, fang2008feedback, mukherjee2012direct} have been widely explored. They transform the non-linear dynamics of the drone into an equivalent linear model so that a suitable flight controller can be designed with the linear control theory. 
However, these controllers often exhibit lack of robustness since the equivalent system may hold zero dynamics (\ie states which are unobservable from system output), causing instability.
Backstepping control approaches are another possible solution \cite{madani2006backstepping,cabecinhas2014nonlinear}. The main advantage is that this family of techniques inherently ensures robustness and prevents the cancellation of valuable non-linearities.
An additional popular alternative that allows to optimize the performance while considering trajectory constraint is represented by non-linear MPC strategies \cite{neunert2016fast, falanga2018pampc}. However, this framework relies heavily on the availability of an accurate model of the quadrotor, an assumption that, in practice, is often difficult to satisfy and therefore it should be learned. Furthermore, it is computationally expensive and may be difficult to be deployed on platforms with low computing capacity. 

\textbf{Learning-based Methods}. For these reasons, learning-based techniques have recently drawn the attention of the robotics community. \changes{In \cite{li2022learning} an iterative learning MPC is proposed to improve task performance over many trials, while in \cite{salzmann2023real} the authors employ deep neural networks in the MPC prediction step to achieve real-time computation on an embedded platform. While these solutions heavily rely on the combination between a module that learns the model dynamics and a classic controller, many recent works are moving toward end-to-end approaches that leverage reinforcement learning to directly learn the control policy from experience.}
This allows to obtain effective control policies that are capable to generalize over complex scenarios without the need for prior knowledge about the non-linear system. For example, the authors in \cite{hwangbo2017control} show a reinforcement learning controller trained to directly map state representations to actuator commands. The authors demonstrate the ability of their solution to stabilize the quadrotor even under extremely challenging conditions, while remarkably reducing computation time by two orders of magnitude if compared to conventional trajectory optimization algorithms. While in \cite{hwangbo2017control} only hovering stabilization is considered, more recent works manage to develop policies able to perform \changesIROS{robust target tracking \cite{dionigi2023exploring},} champion-level drone racing \cite{kaufmann2023champion} and aggressive quadrotor flight \cite{sun2022aggressive}. Further constraints such as the time of flight and the presence of obstacles are taken into consideration in \cite{penicka2022learning}, where a RL neural-network controller \changes{is able to perform} minimum-time quadrotor flight in cluttered environments. 

Despite the promising results shown by RL-based approaches, to foster future research in this direction, it is necessary to perform in-depth studies on the design choices of the RL framework. In this work, we specifically focus on the data input choices since most of the aforementioned works make different choices regarding the sensor data provided as input to the policy network or the output signal that it computes (\eg collective thrust and body rates or single motor thrusts) without analyzing the effect that these choices have on the system performance. This analysis is even more critical if we consider that RL algorithms usually trained in simulation environments cannot be directly optimized on real drone platforms due to battery constraints and safety aspects. Therefore, the design strategy might have a significant impact on the generalization capabilities of the policy network. 

The first benchmark study of different configurations of the policy output is proposed in \cite{kaufmann2022benchmark}, where RL model that computes linear velocities, single-rotor thrusts and collective thrust and body rates are compared. The authors show that the latter strategy is the most robust and effective, particularly with respect to the sim-to-real transfer.
Nonetheless, to the best of our knowledge, the impact of the input configuration on the learned policy has not been investigated in the state of the art. Specifically, the following questions still remain unanswered: \textit{i) What is the minimum information required by the network to compute the output command?} and \textit{ii) What combination of input data provides the best performance?}

\section{Methodology}
\label{sec:method}

\subsection{Quadrotor Dynamic Model}

Training RL methods directly on real drones is impractical 
and unfeasible due to time-consuming and costly real-world experiments, as well as safety concerns regarding potential crashes and hazardous situations. Consequently, the RL agents described in the following are optimized by leveraging a simulator of the quadrotor dynamic model in which the drone is controlled by collective thrust and body rates (CTBR). The output of the policy has been chosen to be CTBR as suggested by the study presented in \cite{kaufmann2022benchmark}. 

Following \cite{mueller2015computationally}, we assume that the quadrotor is a 6 degrees-of-freedom rigid body of mass $m$ and moment of inertia matrix $J = diag(J_x, J_y, J_z)$ in which the evolution of the collective thrust $f$ and the angular velocities $\omega = (\omega_x, \omega_y, \omega_z)$ are modeled as first-order systems with time constant $k_f$ and $k_{\omega}$, respectively. Consequently, the state space is 19-dimensional and the dynamics of the system in the world frame $\mathcal{W}$ can be expressed as follows
\begin{equation}
\label{eq:dyn}
\begin{bmatrix}
\changesIROS{\dot{p}}\\ 
\changesIROS{\dot{v}}\\ 
\dot{R}\\ 
\dot{\omega}\\ 
\dot{f}
\end{bmatrix} = 
\begin{bmatrix}
\changesIROS{v}\\ 
\frac{1}{m}(R_3f+f_{drag}) + g\\ 
R\left [  \omega\right ]_{\times}\\ 
\changesIROS{J^{-1}(k_{\omega}(\omega_{cmd}-\omega)-\left [ \omega \right ]_\times  J \omega )}\\
\changesIROS{k_f(f_{cmd}-f)}
\end{bmatrix},
\end{equation}
where $p$, $v$ and $R$ are the quadrotor absolute position, velocity and \changesIROS{rotation matrix}, respectively. The gravity vector is denoted by $g=\begin{bmatrix} 0 & 0 & -9.81 m/s^2\end{bmatrix}^\top$ while $f_{drag}$ is a linear drag term obtained as $f_{drag} = -\begin{bmatrix}k_{vx}v_x & k_{vy}v_y & k_{vz}v_z \end{bmatrix}^\top$ where $\left ( k_{vx}, k_{vy}, k_{vz} \right)$ are suitable drag coefficients. $R_j$ denotes the $j$-th column of $R$ and $[\omega]_\times$ is the skew-symmetric representation of $\omega$. The input to the system is represented by $f_{cmd}$ and $\omega_{cmd}$, which are respectively the commanded total thrust and body rates. Furthermore, in order to adapt the quadrotor dynamics for the reinforcement learning training process, a zero-order-hold discretization technique is applied. 


\subsection{Problem Formulation}

\changes{The primary focus of this work is to conduct a study to demonstrate the  impact that different input configurations have on end-to-end DRL policies trained in simulation and then deployed on a real quadrotor without any form of fine-tuning. More specifically, we are interested in showing which information is essential to fly, and which leads to better or worse performance.}
\changes{To this aim}, we consider a \changes{point-to-point navigation} task in which the quadrotor starts from a perturbed initial condition, and the goal is to reach a fixed target position $y_r$ as fast as possible, hovering in-place once reached.
The different controllers are optimized with an end-to-end Deep Reinforcement Learning (DRL) strategy to learn a model that directly maps the observation $o(k)$ to the control action $u(k)$. As in standard RL, training is performed by observing the environment rewards $r(k)$ obtained through interactions with the environment across multiple episodes. 

As specified above, we follow \cite{kaufmann2022benchmark} and define a continuous actions space with two signals $\{ f_{cmd}(k), \omega_{cmd}(k) \}$, \ie the collective thrust and the body rates (CTBR). In addition, to reject constant disturbance accelerations, we augment the thrust signal with an integral contribution in the form of $f_{cmd}(k) = f_{cmd}(k-1) + df(k) * ts$, where $ts$ is the sampling time and $u_k = (df(k), \omega_{cmd}(k))$ is the output of the learned policy in which $df(k)$ represents a collective thrust increment.


To study the effect of the different observation space configurations, we design and optimize multiple DRL agents. The considered configurations take inspiration from choices made in several previous works \cite{hwangbo2017control, dionigi2023exploring, kaufmann2023champion, sun2022aggressive, penicka2022learning, kaufmann2022benchmark}. Specifically, the observation processed by each agent is composed of different combinations of the following information: the position error of the quadrotor with respect to the target position $e = y_r - p$, the current rotation matrix $R$ between the body and the world frame, the angular velocities $\omega$ of the drone, and the action $u(k-1)$ computed by the DRL controller at the previous time step. Two reference frames are considered for the position error vector:  the world frame $\mathcal{W}$ and the body frame $\mathcal{B}$. We study also this aspect since absolute or relative position information might be available depending on the onboard sensors. For instance, absolute positioning might be provided by a GPS sensor, while relative information could be obtained through vision devices. 

\begin{table*}[t]
\renewcommand{\arraystretch}{1.3}
\centering
\caption{The considered observation space configurations and their corresponding information.}
\resizebox{\textwidth}{!}{
\label{tab:configuration}
\begin{tabular}{c|c|c|c|c|c|c|c|c}
\hline
\multirow{2}{*}{Input Information} & \multicolumn{8}{c}{Observation Space Configurations}\\
\cline{2-9}
& $\{ \mathbf{e}_\mathbf{\mathcal{W}}, \mathbf{R}, \boldsymbol{\omega},  \mathbf{u}\}$& $\{ \mathbf{e}_\mathbf{\mathcal{W}}, \mathbf{R},  \mathbf{u}\}$& $\{ \mathbf{e}_\mathbf{\mathcal{W}},\boldsymbol{\omega},  \mathbf{u}\}$& $\{ \mathbf{e}_\mathbf{\mathcal{W}},  \mathbf{u}\}$& $\{ \mathbf{e}_\mathbf{\mathcal{B}}, \mathbf{R}, \boldsymbol{\omega},  u\}$& $\{ \mathbf{e}_\mathbf{\mathcal{B}}, \mathbf{R},  \mathbf{u}\}$& $\{ \mathbf{e}_\mathbf{\mathcal{B}}, \boldsymbol{\omega},  \mathbf{u}\}$ & $\{ \mathbf{e}_\mathbf{\mathcal{B}},  \mathbf{u}\}$ \\
\hline
Position Error ($e_\mathcal{W},\;e_\mathcal{B}$)&\cmark&\cmark&\cmark&\cmark&\cmark&\cmark&\cmark&\cmark \\
Rotation Matrix ($R$)&\cmark&\cmark&\xmark&\xmark&\cmark&\cmark&\xmark&\xmark \\
Angular Velocities ($\omega$)&\cmark&\xmark&\cmark&\xmark&\cmark&\xmark&\cmark&\xmark \\
Previous Control Action ($u_{k-1}$)&\cmark&\cmark&\cmark&\cmark&\cmark&\cmark&\cmark&\cmark \\
\hline
\end{tabular}}
\vspace{-1.0em}
\end{table*}


We consider a history of $H=10$ observations to build the input vector of the DRL agent, resulting in $\mathbf{o}(k) = \{o(k),~o(k-1), \cdots,~o(k-H+1)\}$ at the timestep $k$.
Therefore, eight different DRL agents are considered, each one with the observation space configuration reported in Table \ref{tab:configuration}. For brevity, in the following of the paper we refer to the learned models by using their respective observation space such as $\{ \mathbf{e}_\mathbf{\mathcal{W}}, \mathbf{R}, \boldsymbol{\omega},  \mathbf{u}\}$, where the bold is used to indicate that for each information we collect the last $H$ readings.
It follows that the considered hovering problem requires to learn a suitable DRL policy $\pi$ capable of bringing the position error $e$ to zero using the learned control action $u(k) = \pi(\mathbf{o}(k))$ assumed to take values in a continuous action space.


The reward signal $r(k)$ is designed to address the \changes{navigation} problem. Since the main control objective is to reduce the position error to zero, we define the following reward
\begin{equation} \label{rew:track}
    r_{e}(k) = (r_x(k) \, r_y(k) \, r_z(k))^{\beta},
\end{equation}
where
\begin{equation*} \label{rew:axis}
\begin{matrix}


\changes{r_j = \max(0, 1 - \left | e_{j} \right |),}
\end{matrix}
\end{equation*}
\changes{$r_{j}$} and $e_{j}$ are the $j$-th entries of $r_e$ and $e$, respectively, and $\beta>0$ is an appropriate exponent. It should be noted that $r_{e}(k)$ is maximized when the quadrotor reaches the target position.
Furthermore, to also optimize the control effort, we define a penalty $r_{u}$ as
\begin{equation} \label{pen:effort}
    r_{u}(k) = \frac{\| u(k) \|}{1 + \| u(k) \|}.
\end{equation}
To speed up the training process, we provide the DRL agent with a high negative reward when the quadrotor deviates excessively from the desired target point.
Therefore, the overall reward function is
\begin{equation} \label{rew:total}
r(k) = \left\{\begin{array}{l c}
r_{e}(k) \!-\! k_u r_{u}(k) & \| e(k) \| < e_{m} \\[1mm] 
-c, & \text{otherwise}
\end{array}\right.,
\end{equation}
where $k_u>0$ is a weighting parameter that allows a trade-off between the two reward terms, $e_{m}$ is the maximum distance allowed, and $c$ is a large positive constant.

\subsection{Deep Reinforcement Learning Approach}

Since in the quadrotor setting, both the observation and the output spaces are continuous, we leverage the popular policy gradient paradigm to design the DRL agents. Specifically, we exploit the \textit{asymmetric actor-critic} framework \cite{pinto2017asymmetric, dionigi2022vat} and use Deep Neural Networks (DNNs) approximators to implement the two distinct actor and critic architectures. More specifically, the \textit{actor} (A-DNN) learns the optimal control policy $\pi(\mathbf{o}(k))$ while the \textit{critic} (C-DNN) is responsible for evaluating such a policy at training time.

The A-DNN is a Multi-Layer Perceptron (MLP) with three hidden layers, each one composed of 256 neurons and $\tanh$ activations. The network input depends on the specific model considered in the benchmark study and it is obtained by flattening $\mathbf{o}(k)$ in a $N * H$-dimensional vector, where $N$ is the length of each  $o(i)$, with $i=k, \dots, k-H+1$ in the sequence (\eg for $\{ \mathbf{e}_\mathbf{\mathcal{W}}, \mathbf{R}, \boldsymbol{\omega},  \mathbf{u}\}$ we have $N=19$). Every agent shares instead the output configuration, composed of the four-dimensional vector $u(k) = \{df(k), \omega_{cmd}(k)\}$ representing the control commands to the drone.
Furthermore, to alleviate the steady-error problem derived by the critic overestimation bias \cite{kuznetsov2020controlling}, we impose $u(k) = u(\mathbf{o}(k)) - u(\mathbf{o}_0)$ since in perfect hovering conditions, \ie when the quadrotor has reached the target point, the output of the network should be zero ($\mathbf{o}_0$ is the observation in perfect hovering conditions).

Thanks to the asymmetric framework, we can provide the C-DNN with more privileged information since it operates exclusively during the training phase. In particular, the input to the critic network is augmented with instantaneous velocities and accelerations to facilitate training, \ie $\left [ p\; \dot{p}\; \ddot{p}\; R\; \omega \right ]$. The C-DNN architecture is composed of a MLP with three hidden layers, each one with 256 neurons and ReLu activations. The final layer of the network estimates the action-value $Q_\mathbf{\pi}$. It is important to remark that the critic used to optimize each DRL agent has the same network structure and inputs.

\begin{table*}[h]
\renewcommand{\arraystretch}{1.3}
\centering
\caption{Experimental results: positional tracking error in centimeters across different scenarios.}
\resizebox{\textwidth}{!}{
\label{tab:results}
\begin{tabular}{|c|c|c|c|c|c|c|c|c|c|c|c|c|c|c|c|c|c|c|c|c|c|c|c|c|c|c|c|c|}
\hline
\multirow{3}{*}{\shortstack{Observation\\Space\\Configuration}} & \multicolumn{28}{c|}{Experimental Scenarios and Metrics}\\
\cline{2-29}
& \multicolumn{4}{c|}{Hovering} & \multicolumn{4}{c|}{Planar Ellipse} & \multicolumn{4}{c|}{Planar Eight-Shape} & \multicolumn{4}{c|}{3D Eight-Shape} & \multicolumn{4}{c|}{Planar Ellipse - Inc. Speed} & \multicolumn{4}{c|}{Planar Ellipse - Inc. Speed} & \multicolumn{4}{c|}{Planar Ellipse - Inc. Speed}\\
\cline{2-29}
& $P_{x}$ & $P_{y}$ & $P_{z}$ & $P_{c}$ & $P_{x}$ & $P_{y}$ & $P_{z}$  & $P_{c}$ & $P_{x}$ & $P_{y}$ & $P_{z}$  & $P_{c}$ & $P_{x}$ & $P_{y}$ & $P_{z}$ & $P_{c}$ & $P_{x}$ & $P_{y}$ & $P_{z}$ & $P_{c}$ & $P_{x}$ & $P_{y}$ & $P_{z}$ & $P_{c}$ & $P_{x}$ & $P_{y}$ & $P_{z}$ & $P_{c}$\\
\hline
$\{ \mathbf{e}_\mathbf{\mathcal{W}}, \mathbf{R}, \boldsymbol{\omega},  \mathbf{u}\}$ & $  5.0$ & $  6.9$ & $  \textbf{2.1} $ & $ \cellcolor{red!30} 4.6$ & $  7.9$ & $  6.6 $ & $ \textbf{2.7} $ & $ \cellcolor{red!30} 5.7$ & $ 9.8$ & $  7.1 $ & $  3.4 $ & $ \cellcolor{red!30} 6.8$ & $  9.6$ & $  8.5 $ & $  6.0 $ & $ \cellcolor{red!30} 8.1$ & $  18.3$ & $  10.4$& $  3.5 $& $ \cellcolor{red!30} 10.7$& $  13.8 $& $  10.4 $& $  5.0 $& $ \cellcolor{red!30} 9.7$& $  15.0 $& $ 10.3$& $  6.9$& $ \cellcolor{red!30} 10.7$ \\
\hline
$\{ \mathbf{e}_\mathbf{\mathcal{W}}, \mathbf{R},  \mathbf{u}\}$ & $  2.6 $ & $  \textbf{2.1} $ & $  3.3$ & $ \cellcolor{green!30} \textbf{2.7}$ & $  \textbf{2.3} $ & $  3.4 $ & $  3.2$ & $ \cellcolor{green!30} \textbf{3.0}$ & $  \textbf{2.7} $ & $  \textbf{2.8} $ & $ 4.1 $ & $ \cellcolor{green!30} \textbf{3.1}$ & $  \textbf{2.5}$ & $ 2.9 $ & $  6.2$ & $ \cellcolor{green!30} \textbf{3.9}$ & $  \textbf{3.6} $ & $  10.9 $& $  3.8 $& $ \cellcolor{green!30} \textbf{6.1}$& $ \textbf{4.0}$& $  \textbf{3.6} $& $  4.3 $& $ \cellcolor{green!30} \textbf{4.0}$& $  \textbf{3.8} $& $  \textbf{3.9} $& $  5.7$& $ \cellcolor{green!30} \textbf{4.5}$\\
\hline
$\{ \mathbf{e}_\mathbf{\mathcal{B}}, \mathbf{R}, \boldsymbol{\omega},  \mathbf{u}\}$ & $  4.7 $ & $  3.0 $ & $  3.2 $ & $ \cellcolor{red!30} 3.7$ & $  6.7 $ & $  4.2 $ & $  4.4 $ & $ \cellcolor{red!30} 5.1$ & $  9.5 $ & $  3.1 $ & $  \textbf{3.1} $ & $ \cellcolor{red!30} 5.2$ & $  8.0 $ & $  2.9$ & $  6.6 $ & $ \cellcolor{red!30} 5.8$ & $  7.2 $ & $  15.6 $& $  9.2 $& $ \cellcolor{red!30} 10.7$& $  9.9 $& $  4.4 $& $  4.5$& $ \cellcolor{red!30} 6.3$& $  10.3$& $  4.7 $& $  7.5$& $ \cellcolor{red!30} 7.5$\\
\hline
$\{ \mathbf{e}_\mathbf{\mathcal{B}}, \mathbf{R},  \mathbf{u}\}$ & $  2.5 $ & $ 2.7 $ & $  3.1 $ & $ \cellcolor{green!30} \textbf{2.7}$ & $  4.0 $ & $  \textbf{3.3} $ & $ 3.6$ & $ \cellcolor{green!30} 3.6$ & $  4.1 $ & $3.0 $ & $  3.6 $ & $ \cellcolor{green!30} 3.5$ & $  3.1 $ & $  \textbf{2.6} $ & $ \textbf{5.8} $ & $ \cellcolor{green!30} \textbf{3.9}$ & $  7.7 $ & $  11.8 $& $  7.0 $& $ \cellcolor{yellow!30} 8.8$& $  7.0 $& $  4.9$& $  4.2$& $ \cellcolor{yellow!30} 5.4$& $  8.5$& $  6.1 $& $  6.8 $& $ \cellcolor{yellow!30} 7.1$\\
\hline
$\{ \mathbf{e}_\mathbf{\mathcal{B}}, \boldsymbol{\omega},  \mathbf{u}\}$ & $  \textbf{2.1} $ & $  3.2 $ & $  2.7 $ & $ \cellcolor{green!30} \textbf{2.7}$ & $  3.7 $ & $  5.0 $ & $  3.2$ & $ \cellcolor{yellow!30} 4.0$ & $  3.5$ & $  4.2 $  & $ 3.3 $ & $ \cellcolor{yellow!30} 3.6$ & $  4.2 $ & $  4.5 $ & $  \textbf{5.8} $ & $ \cellcolor{yellow!30} 4.8$ & $  9.1 $ & $  \textbf{9.6} $& $  4.0 $& $ \cellcolor{yellow!30} 7.6$& $  4.4 $& $  5.7 $& $  \textbf{3.1} $& $ \cellcolor{green!30} 4.3$& $  5.5 $& $  6.4 $& $  5.9 $& $ \cellcolor{yellow!30} 5.9$\\
\hline
$\{ \mathbf{e}_\mathbf{\mathcal{B}},  \mathbf{u}\}$ & $ 2.2$ & $  3.7$ & $  3.1$ & $ \cellcolor{yellow!30} 3.0$ & $  3.2$ & $  4.7$ & $  3.0$ & $ \cellcolor{green!30} 3.6$ & $  3.3$ & $  4.2 $  & $  3.9 $ & $ \cellcolor{yellow!30} 3.8$ & $  3.5$ & $  4.3$ & $ 6.7$ & $ \cellcolor{yellow!30} 4.8$ & $  7.3 $ & $  10.1 $& $  \textbf{2.9} $& $ \cellcolor{green!30} 6.8$& $  7.0 $& $  7.2 $& $  3.3 $& $ \cellcolor{yellow!30} 5.9$& $  5.0 $& $  6.9 $& $  \textbf{4.7} $& $ \cellcolor{green!30} 5.5$\\
\hline
\hline
PID Controller & $ 1.6$ & $  1.5$ & $  1.4$ & $ \cellcolor{green!30} 1.5$ & $  2.4$ & $  5.5$ & $  1.6$ & $ \cellcolor{green!30} 3.2$ & $  2.8$ & $  2.7 $  & $  1.5 $ & $ \cellcolor{green!30} 2.3$ & $  3.4$ & $  2.6$ & $ 8.7$ & $ \cellcolor{yellow!30} 4.9$ & $  7.8 $ & $  15.6 $& $  2.2 $& $ \cellcolor{yellow!30} 8.5$& $  7.9 $& $  5.8 $& $  1.5 $& $ \cellcolor{yellow!30} 5.1$& $  8.2 $& $  5.8 $& $  16.8 $& $ \cellcolor{red!30} 10.3$\\
\hline
\end{tabular}}
\vspace{-1.5em}
\end{table*}

\subsection{Training and Implementation Details}

In order to train the DRL agents we employ the Soft Actor-Critic (SAC) \cite{haarnoja2018soft} framework\changes{, which is one of the most recent RL strategies in literature. SAC often exhibits more stable training dynamics due to its use of entropy regularization, and, thanks to its off-policy nature, can take advantage of a replay buffer that incorporates experience also from past episodes, which, in general, allows for more effective generalization capabilities.}
\changes{Moreover,} we use a domain randomization strategy \cite{kaufmann2022benchmark, dionigi2023exploring} to achieve zero shot sim-to-real transfer and improve robustness with respect to model uncertainties. In particular, we randomize the mass $m$ and the inertia matrix $J$ of the drone, the random bias added to the gravity vector $g$, and the drag linear coefficients $\left ( k_{vx}, k_{vy}, k_{vz} \right)$ \changesIROS{up to $\pm 10\%$ of their nominal values}. Furthermore, in order to favor robustness to the control lags on real platforms, we add random delays to the control actions computed by the neural network \changesIROS{in the interval $\left[0;10\right]ms$}.

Each DRL agent is optimized through the Stable-Baselines~\cite{stable-baselines3} implementation of SAC, which we customized to the asymmetric actor-critic framework. 
The networks of each model are trained for a total of about $800k$ episodes with $8$ parallel environments. We use the Adam optimizer with a learning rate of $0.0003$ and a batch size of $256$.

The training process is structured in episodes. At the beginning of each one of them, the quadrotor is randomly positioned inside the environment distant from the desired target goal point. The drone starts from a perturbed initial condition with random orientation, linear and angular velocities (different from the hovering ones). Hence, during a training episode the DRL agent has to perform control actions in order to (i) recover the quadrotor from the initialization state, (ii) guide it to the desired target point, and (iii) hover in-place once arrived at destination. An episode terminates either if the step number $k$ reaches a predefined maximum limit or the distance limit from the target point $\| e(k) \| > e_{m}$ is violated.

The training process of each model requires about $2$ hours and $1.1$ GB of VRAM to converge on a workstation equipped with $2$ × NVIDIA RTX 2080Ti with 11GB of VRAM, an Intel Core processor i7-9800X ($3.80$ GHz$\times 16$) and $64$ GB of DDR4 RAM. The inference time required for a control action is about $3$ ms on a NVIDIA Jetson Xavier NX.

\section{Experimental Results}
\label{sec:experiments}

\subsection{Experimental Setup and Metrics}
To quantitatively evaluate the impact of each observation configuration, each DRL agent trained in simulation is deployed on a real platform without further optimization procedure, \ie with zero-shot adaptation. We perform an extensive experimental campaign and evaluate the performance of each model with respect to \changes{set-point stabilization and} tracking of different trajectories at varying velocities. Hovering experiments are designed to compare how different input configurations affect the drone stabilization capabilities. In addition, three additional trajectories, \ie an ellipse and two eight-shape trajectories, one planar and one that spans the 3D space, are also considered (refer to the supplementary multimedia material). Trajectory tracking experiments are designed to evaluate the robustness and the capabilities of the model to respond to set point variations across time. In addition, to explore the responsiveness of the model, we double the average speeds for each trajectory type trial, \changes{we refer to these experiments as ``Increased Speed" (Inc. Speed)}.

In each trial, the robot executes a specific trajectory for approximately $20$ seconds. To compare the models and extract quantitative results from the experiments we employ the following metrics
\begin{equation*}
\begin{matrix}
    \tilde{P}_{j}(k) = \sqrt{\sum_{k=0}^{M}\frac{(y_j(k) - p_j(k))^2}{M}},
\end{matrix}
\end{equation*}
\vspace{-1em}
\begin{equation*}
\begin{matrix}
    \tilde{P}_{c}(k) = \frac{\tilde{P}_{x}(k) + \tilde{P}_{y}(k) + \tilde{P}_{z}(k)}{3},
\end{matrix}
\end{equation*}
where $\tilde{P}_{j}(k)$ is the Root Mean Square Error (RMSE) along the trajectory of the quadrotor position with respect to the three Cartesian $j$ axis, and $\tilde{P}_{c}(k)$ is an overall trajectory tracking score obtained by averaging the $\tilde{P}_{j}(k)$.
Each experiment is repeated three times and the RMSEs are averaged over both the episode time and the runs performed in each scenario, resulting in
\begin{equation*}
	P_{m} = \frac{1}{3 N_c} \sum_{i=1}^{3} \sum_{k=0}^{N_c-1} {}^{(i)\!}\tilde{P}_{m}(k),
\end{equation*}
where $m \in \{x, y, x,  c\}$, ${}^{(i)\!}\tilde{P}$ indicates that the performance is evaluated on the $i$-th run, and $N_c$ is the number of samples within the episode. 


\subsection{Discussion}

In Table \ref{tab:results}, we report the results of the experimental campaign. A first key finding of this study is that the majority of the models successfully learn effective policies for controlling the quadrotor in real-world scenarios, even those with a less informative observation space. 
Only two DRL agents do not reach convergence, \ie those with the $\{ \mathbf{e}_\mathbf{\mathcal{W}}, \boldsymbol{\omega},  \mathbf{u}\}$ and $\{ \mathbf{e}_\mathbf{\mathcal{W}}, \mathbf{u}\}$ observation space configurations (hence, we do not report them in Table \ref{tab:results}). This result is expected since the position error $\mathbf{e}_\mathbf{\mathcal{W}}$ is given in the world reference frame $\mathcal{W}$ while the action space of the DRL controller (CTBR) is expressed in the body frame $\mathcal{B}$. 
Therefore, the information on the rotation matrix is crucial to recover the orientation of the drone and control it. 

Notably, our intuition that providing more information to the agent does not necessarily lead to better performance is supported by these results. By observing the values of the $P_c$ metric in the Table we notice that the models $\{ \mathbf{e}_\mathbf{\mathcal{W}}, \mathbf{R}, \boldsymbol{\omega},  \mathbf{u}\}$ and $\{ \mathbf{e}_\mathbf{\mathcal{B}}, \mathbf{R}, \boldsymbol{\omega},  \mathbf{u}\}$ achieve the worst performance. 
While this might be surprising, it confirms that requiring the agent to process more information increases the complexity of the model and might hinder the optimization process. In DRL settings, the experience needed to learn the optimal policy grows significantly as the observation space increases and, as in the case of these two models, could negatively impact the training convergence. 

Better scores are achieved by $\{ \mathbf{e}_\mathbf{\mathcal{W}}, \mathbf{R},  \mathbf{u}\}$ and $\{ \mathbf{e}_\mathbf{\mathcal{B}}, \mathbf{R}, \mathbf{u}\}$. The former shows remarkable hovering and trajectory tracking capabilities in every considered scenario, while we observe a performance degradation for the $\{ \mathbf{e}_\mathbf{\mathcal{B}}, \mathbf{R}, \mathbf{u}\}$ agent when the velocity increase. Therefore, $\{ \mathbf{e}_\mathbf{\mathcal{W}}, \mathbf{R}, \mathbf{u}\}$ is the observation space configuration that attains the best performance. It should be noted that this input configuration contains the minimum required information when position error is expressed in the world frame.

It is also very interesting to observe that, despite being the configurations with less information, the $\{ \mathbf{e}_\mathbf{\mathcal{B}}, \boldsymbol{\omega},  \mathbf{u}\}$ and $\{ \mathbf{e}_\mathbf{\mathcal{B}},  \mathbf{u}\}$ agents are able to learn robust flying policies and achieve comparable performance with respect to $\{ \mathbf{e}_\mathbf{\mathcal{W}}, \mathbf{R},  \mathbf{u}\}$. This is remarkable if we consider that they do not have access to the drone attitude. Moreover, the $\{ \mathbf{e}_\mathbf{\mathcal{B}},  \mathbf{u}\}$ model shows only a little performance drop when higher velocities are considered and achieves better scores than the other model with position errors expressed in the body frame. This result is significant since it proves that it is possible to fly and perform trajectory tracking by utilizing only relative position information, which can be obtained from low-cost sensors such as RGB cameras or Ultra-Wide Band (UWB) devices.

\changes{Furthermore, to better position the DRL controllers of this benchmark study with respect to the classic control methods, we add a comparison against a non-RL baseline. More specifically, we implement a standard quadrotor control architecture featuring two PID feedback loops: an outer loop for position control and an inner one for attitude stabilization \cite{loianno2016estimation, lopez2023pid}. The PID parameters have been tuned experimentally to achieve a suitable trade-off between responsiveness to trajectory tracking errors and sensitivity to noise. As shown by the results in Table \ref{tab:results}, every DRL controller exhibits similar scores with respect to the PID counterpart, without requiring manual parameter tuning.
Moreover, the best performing DRL policies such as $\{ \mathbf{e}_\mathbf{\mathcal{W}}, \mathbf{R},  \mathbf{u}\}$, $\{ \mathbf{e}_\mathbf{\mathcal{B}}, \boldsymbol{\omega},  \mathbf{u}\}$ and $\{ \mathbf{e}_\mathbf{\mathcal{B}},  \mathbf{u}\}$ demonstrate less degradation in performance while tracking faster trajectories and achieve higher metric scores with respect to the PID.} 

\changesIROS{Nevertheless, even if better results are achieved against the PID baseline, it is important to highlight that this comparison is for reference purpose only. As also shown in \cite{kaufmann2022benchmark}, there are more complex controllers, such as a non-linear MPC, capable of achieving sub-centimeter trajectory tracking errors. However, the objective of the paper is not to propose a novel controller, but to analyze the sim-to-real adaptation capabilities of DRL policies with respect to the observation space.}

\subsection{\changes{Velocity Robustness}}
\changes{We include a stress experiment to understand the maximum velocity that each DRL controller is able to track before a failure. To this aim, we move the target point across a circular trajectory and we gradually increase the speed until the quadrotor loses the tracking (\ie exceeds 50 cm of distance from the target point). In Table \ref{tab:velfail}, we report the corresponding maximum supported velocity until a failure for each DRL controller. It is important to highlight that such analysis can be performed only on a simulation environment, since pushing the controller to the limit of stability can damage the real platform and pose security issues. The numerical results reveal that $\{ \mathbf{e}_\mathbf{\mathcal{W}}, \mathbf{R}, \boldsymbol{\omega},  \mathbf{u}\}$ and $\{ \mathbf{e}_\mathbf{\mathcal{B}}, \mathbf{R}, \boldsymbol{\omega},  \mathbf{u}\}$ exhibit less robustness to fast trajectories, despite having access to a more privileged observation space compared to the other models. On the other hand, $\{ \mathbf{e}_\mathbf{\mathcal{B}}, \boldsymbol{\omega}, \mathbf{u}\}$ and $\{ \mathbf{e}_\mathbf{\mathcal{B}},  \mathbf{u}\}$ are capable of tracking a target point moving at approximately 1.2 $m/s$, while $\{ \mathbf{e}_\mathbf{\mathcal{W}}, \mathbf{R},  \mathbf{u}\}$ and $\{ \mathbf{e}_\mathbf{\mathcal{B}}, \mathbf{R},  \mathbf{u}\}$ also demonstrate remarkable performance with a maximum velocity of about 1.0 $m/s$. These results further strengthen the generalization capabilities of the DRL agents, since we trained the controllers on a different task than fast trajectory tracking.}

\begin{table}[h]
\renewcommand{\arraystretch}{1.3}
\centering
\caption{Velocity stress experiment results ($m/s$).}
\resizebox{\columnwidth}{!}{
\label{tab:velfail}
\begin{tabular}{c|c|c|c|c|c}
\hline
$\{ \mathbf{e}_\mathbf{\mathcal{W}}, \mathbf{R}, \mathbf{\omega},  \mathbf{u}\}$ & $\{ \mathbf{e}_\mathbf{\mathcal{W}}, \mathbf{R},  \mathbf{u}\}$ & $\{ \mathbf{e}_\mathbf{\mathcal{B}}, \mathbf{R}, \mathbf{\omega},  \mathbf{u}\}$ & $\{ \mathbf{e}_\mathbf{\mathcal{B}}, \mathbf{R},  \mathbf{u}\}$ & $\{ \mathbf{e}_\mathbf{\mathcal{B}}, \mathbf{\omega},  \mathbf{u}\}$ & $\{ \mathbf{e}_\mathbf{\mathcal{B}},  \mathbf{u}\}$ \\
\hline
0.73 & 1.06 & 0.74 & 0.98 & 1.21 & 1.19 \\
\hline
\end{tabular}}
\end{table}

\subsection{\changes{Ablation Study}}

\changes{In order to justify two important design choices of this benchmark, we conduct an ablation study regarding (i) the use of a $H$-length window for the observation space and (ii) the selection of the proposed input configurations.}

\begin{figure}[t]
    \centering
    \includegraphics[width=\columnwidth]{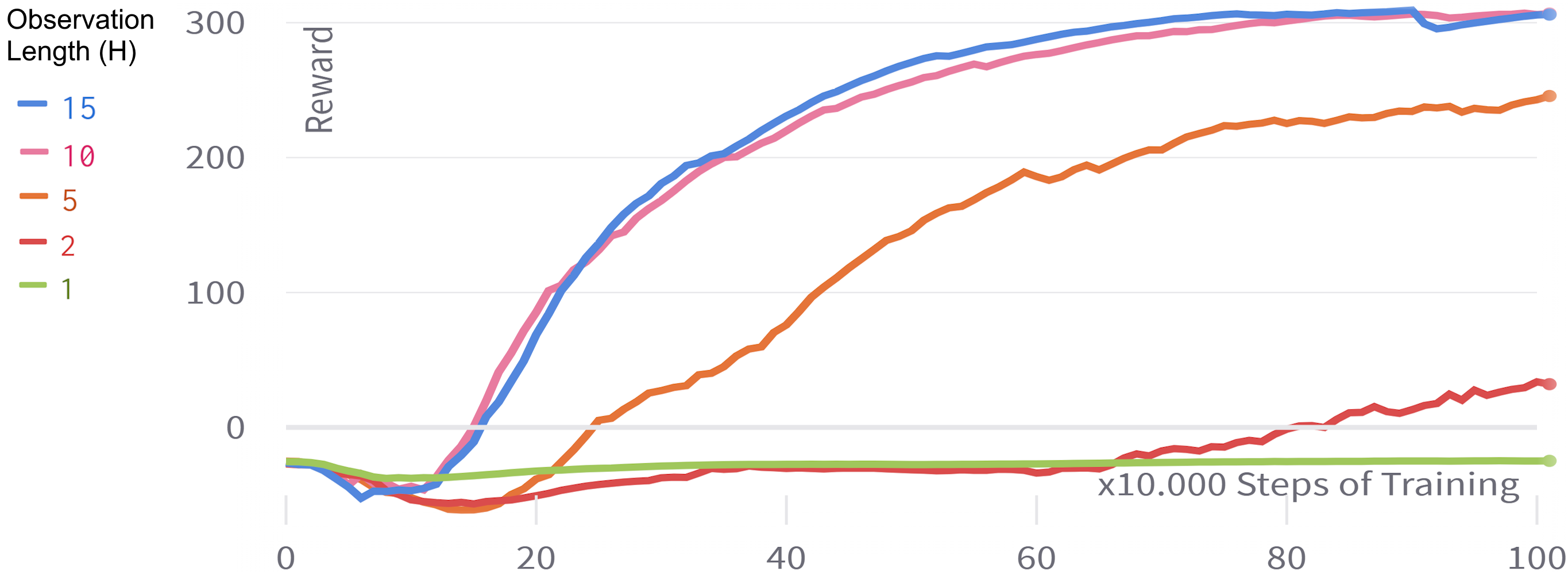}
    \caption{\changes{Learning curves of the  $\{ \mathbf{e}_\mathbf{\mathcal{W}}, \mathbf{R},  \mathbf{u}\}$ variants trained with a different observation length: we report the cumulative reward reached by each agent during the training phase.}}
    \vspace{-1.5em}
    \label{fig:window} 
\end{figure}

\noindent \changes{\textbf{Observation Length} ($H$):} 
\changes{We take into consideration the best-performing controller of the benchmark, $\{ \mathbf{e}_\mathbf{\mathcal{W}}, \mathbf{R},  \mathbf{u}\}$, and we trained four variants by changing the length of the window within the set $\{1, 2, 5, 10, 15\}$. As shown by the plot in Figure \ref{fig:window}, the use of a single observation does not lead to convergence. However, as the length of the observation window increases, the learning process becomes more effective. We observe performance saturation with a window length of $10$, which aligns with findings from other state-of-the-art works \cite{kaufmann2022benchmark}. Consequently,
we fix the observation window to $10$, ensuring the controller's effectiveness while avoiding unnecessary complexity in the neural network architecture. }

\noindent \changes{\textbf{Observation Configuration}:} 
\changes{In order to choose the policy input configurations, we conducted an extensive preliminary study in which a more comprehensive set of possible information was considered, resulting in our decision to not include the velocity and the quaternion. To prove the effectiveness of our choices, we train two variants of the $\{ \mathbf{e}_\mathbf{\mathcal{W}}, \mathbf{R},  \mathbf{u}\}$ model. In the former, called $\{ \mathbf{e}_\mathbf{\mathcal{W}}, \mathbf{v}_\mathbf{\mathcal{W}},\mathbf{R},  \mathbf{u}\}$, we augment the observation space with the quadrotor linear velocity w.r.t. the world frame $\mathcal{W}$, while in the latter, named $\{ \mathbf{e}_\mathbf{\mathcal{W}}, \mathbf{q},  \mathbf{u}\}$, we replace the rotation matrix with the corresponding quaternion (q). As shown by the numerical results reported in Table \ref{tab:ablvelquat}, the velocity does not improve the performance of the DRL controller, and the rotation matrix is the most effective representation of the orientation for the considered task. In fact, (i) the velocity can be inferred from a sequence of observations, and consequently, without loss of information, we prefer to not consider it in the benchmark, and (ii) the rotation matrix is a non-ambiguous representation in the space, while the quaternion is not (\eg q is equal to $-$q), and consequently, we preferred the former since using the latter can lead to unexpected behaviors at test time.}

\vspace{-0.5em}
\begin{table}[h]
\renewcommand{\arraystretch}{1.3}
\centering
\caption{Experiments on input configurations ($cm$).}
\resizebox{\columnwidth}{!}{
\label{tab:ablvelquat}
\begin{tabular}{|c|c|c|c|c|c|c|c|c|c|c|c|c|c|c|c|c|}
\hline
\multirow{3}{*}{\shortstack{Observation\\Space\\Configuration}} & \multicolumn{12}{c|}{Experimental Scenarios and Metrics}\\
\cline{2-13} & \multicolumn{4}{c|}{Planar Ellipse} & \multicolumn{4}{c|}{Planar Eight-Shape} & \multicolumn{4}{c|}{3D Eight-Shape}\\
\cline{2-13}
& $P_{x}$ & $P_{y}$ & $P_{z}$ & $P_{c}$ & $P_{x}$ & $P_{y}$ & $P_{z}$  & $P_{c}$ & $P_{x}$ & $P_{y}$ & $P_{z}$  & $P_{c}$ \\
\hline
$\{ \mathbf{e}_\mathbf{\mathcal{W}}, \mathbf{R},  \mathbf{u}\}$ & $  \textbf{2.7} $ & $  11.4 $& $  2.3 $& $ \cellcolor{green!30} \textbf{5.5}$& $ \textbf{3.3}$& $  2.6 $& $  \textbf{1.5} $& $ \cellcolor{green!30} \textbf{2.5}$& $  \textbf{2.8} $& $  2.8 $& $  4.0$& $ \cellcolor{green!30} \textbf{3.2}$ \\
\hline
$\{ \mathbf{e}_\mathbf{\mathcal{W}}, \mathbf{v}_\mathbf{\mathcal{W}}, \mathbf{R},  \mathbf{u}\}$ & $  13.2 $ & $  \textbf{5.5} $& $  \textbf{1.4} $& $ \cellcolor{red!30} 6.7$& $  13.8 $& $  \textbf{1.2} $& $  1.9 $& $ \cellcolor{red!30} 5.6$& $  15.0 $& $  \textbf{1.5} $& $  \textbf{2.8} $& $ \cellcolor{red!30} 6.4$ \\
\hline
$\{ \mathbf{e}_\mathbf{\mathcal{W}}, \mathbf{q},  \mathbf{u}\}$  & $  5.4 $ & $  10.0 $& $  4.4 $& $ \cellcolor{yellow!30} 6.6$& $  4.9 $& $  2.1 $& $  2.7 $& $ \cellcolor{yellow!30} 3.2$& $  5.3 $& $  2.4 $& $  6.1 $& $ \cellcolor{yellow!30} 4.6$ \\
\hline
\end{tabular}}
\vspace{-1.2em}
\end{table}


\section{Conclusion}
\label{sec:conclusion}



\changes{In this work, we presented a benchmark study on how different observation space configurations affect the performance of DRL-based controllers when deployed with zero-shot adaptation on a real quadrotor platform. Several experiments have been performed and, to the best of our knowledge, this is the first study that discusses these fundamental aspects of DRL controller design for MAVs. We believe that this work could become an important reference for future research 
by providing a guideline for selecting the optimal observation space. More specifically, the results we presented suggest that $\{ \mathbf{e}_\mathbf{\mathcal{W}}, \mathbf{R},  \mathbf{u}\}$ and $\{ \mathbf{e}_\mathbf{\mathcal{B}},  \mathbf{u}\}$ provide the best point-to-point navigation performance. Moreover, since the ability to fly is strictly needed in each task involving quadrotors, these configurations can also be considered as reasonable starting points for other applications.} 

\changes{Future works will focus on extending this benchmark by considering agile flight maneuvers as well as other classes of aerial robots such as fixed-wing configurations.}

\balance

\bibliographystyle{IEEEtran}
\bibliography{example}

\end{document}